\documentclass[runningheads]{llncs}
\usepackage{graphicx}
\usepackage{times}

\usepackage{soul}
\usepackage{url}
\usepackage{color}
\usepackage[hidelinks,breaklinks]{hyperref}
\usepackage[utf8]{inputenc}
\usepackage[small]{caption}
\usepackage{amsmath,amssymb}
\usepackage{booktabs}
\usepackage{subfigure}
\usepackage{array}
\usepackage{graphicx}
\usepackage{multirow}
\usepackage{bm}
\usepackage{breakcites}

\begin{document}
\title{Regularized Pooling}
%
%
\author{Takato Otsuzuki\inst{1} \and
Hideaki Hayashi\inst{1}\orcidID{0000-0002-4800-1761} \and
Yuchen Zheng\inst{1}\orcidID{0000-0003-3093-6929} \and
Seiichi Uchida\inst{1}\orcidID{0000-0001-8592-7566}}
\authorrunning{T. Otsuzuki et al.}
%
\institute{Kyushu University, Fukuoka, Japan}
\maketitle              
\begin{abstract}
In convolutional neural networks (CNNs), pooling operations play important roles such as dimensionality reduction and deformation compensation. In general, max pooling, which is the most widely used operation for local pooling, is performed independently for each kernel. However, the deformation may be spatially smooth over the neighboring kernels. This means that max pooling is too flexible to compensate for actual deformations. In other words, its excessive flexibility risks canceling the essential spatial differences between classes. In this paper, we propose \textit{regularized pooling}, which enables the value selection direction in the pooling operation to be spatially smooth across adjacent kernels so as to compensate only for actual deformations. The results of experiments on handwritten character images and texture images showed that regularized pooling not only improves recognition accuracy but also accelerates the convergence of learning compared with conventional pooling operations.


\end{abstract}

\section{Introduction}
\begin{figure}[!t]
	\centering
	\includegraphics[width=0.7\textwidth]{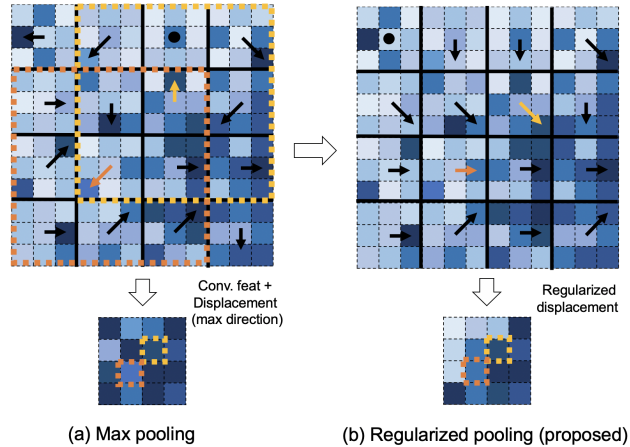}\\[-2mm]
	\caption{(a)~Max pooling and (b)~our regularized pooling. A darker color indicates a larger feature value. The arrow indicates the relative position of the selected value from the center of the kernel. In (a), the maximum value is determined at each kernel. In (b), the value selections in adjacent kernels are regularized (spatially smoothed) and, consequently, a non-maximum value can be selected. The dotted squares indicate the $3\times 3$ window (i.e., $w=3$) for smoothing the direction of selection. \textbf{Although the stride length is identical to kernel size $n=3$ in this figure for a simpler illustration, our method can be realized in arbitrary conditions.}}
	\label{fig:displacementfeature}
\end{figure}

Max pooling in convolutional neural networks (CNNs) is the operation used to select the maximum value in each kernel, as shown in Fig.~\ref{fig:displacementfeature}(a). It plays several important roles in CNN-based image recognition. One is the dimensionality reduction of convolutional features; by using a max pooling operation with an appropriate stride length, we can reduce the size of the convolutional feature map and expect efficient computation as well as information aggregation. Another role is deformation compensation. Even if the convolutional features undergo local (i.e., small) spatial translations due to deformations in the input images, the reduced feature maps are invariant to the translations. Consequently, the CNN becomes robust to deformations in the input images.\par

\begin{figure}[t]
	\centering
	\includegraphics[width=0.8\hsize]{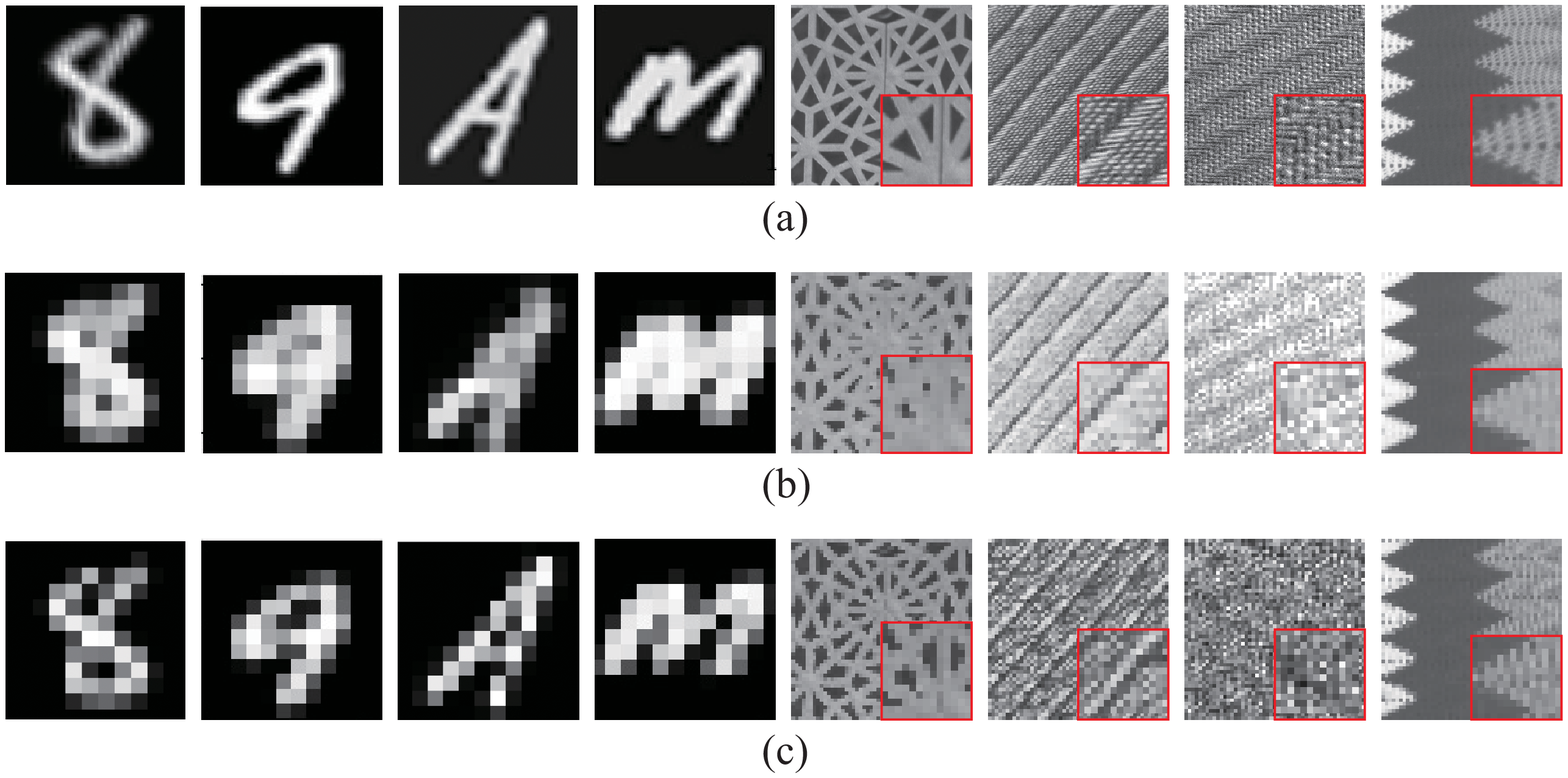}\\[-2mm]
	\caption{(a)~Original image and its results for (b)~max pooling and (c)~our regularized pooling. 
	Both pooling operations are applied to a channel of the first convolution layer with the kernel size $n=5$ and the stride $s=5$. The window size $w$ for the regularization is $3$.}
	\label{fig:pooling-effect}\vspace{-3mm}
\end{figure}

This paper is motivated by the fact that the deformation compensation ability of the max pooling operation is excessive for actual deformations.
Most of the actual deformations are topology-preserving, i.e., spatially continuous within each object region; if a part of an object shifts to a certain direction, its neighboring part also shifts to a similar direction. However, since the value selection by max pooling is performed for each kernel independently, it not only compensates for intra-class topology-preserving deformations, but may also ``over-compensate'' for essential differences between similar classes.\par
Fig.~\ref{fig:displacementfeature}(a) illustrates the excessive flexibility of max pooling, where the kernel size $n$ and the stride $s$ are equally set at 3 for a simpler illustration. (In the later experiments, $s$ was often smaller than $n$.) The arrow on the convolutional feature map shows directions of the maximum value (depicted as the darkest point) from the center of the kernel. The value selection is performed independently at each kernel and therefore the arrows point random directions in the map. If we consider that each arrow represents a local displacement, the arrows work as spatial warping to compensate for the deformations in the map. We can thus understand that these random directions do not fit to continuous deformations. In other words, the flexibility of the max pooling operation is excessive for the actual deformations.\par
The lower part of Fig.~\ref{fig:displacementfeature}(a) shows the result of the max pooling operation. Due to the greedy selection of the maximum value at each kernel, the result of pooling consists mostly of large feature values (i.e., darker colors). However, the original convolutional feature map is not always large; it exhibits a trend that the upper-left side has smaller values and the lower-right side has larger values. The result of max pooling no longer exhibits this trend. This means that max pooling easily overlooks small-valued but important parts and thus might ignore essential differences between similar classes. 
\par
Figs.~\ref{fig:pooling-effect}(a) and (b) show digit and texture images and their results when the max pooling operation is applied to their first convolutional feature map, respectively. The digit images have one or two holes (`8,' `9,' and `A') or near-hole concave parts (`m'). These hole parts have smaller values and thus nearly disappear by the max pooling operation despite their importance for their discrimination. For example, the max pooling result of `8' might be confused with that of `5' (In fact, this `8' is misrecognized as `5' by a CNN with the max pooling operation). Texture images are composed of coarse and fine structures. Their fine structures are lost by max pooling, whereas the coarse ones are still preserved.
\par
In this paper, we propose a {\em regularized pooling} operation, where the flexibility of the pooling operation is regularized to fit the characteristics of actual deformations. Fig.~\ref{fig:displacementfeature}(b) illustrates the proposed regularized pooling operation. The key idea is to smooth the value selection directions in the pooling operation. By taking the average of max value directions in the neighboring kernels in a window (the dotted squares in (a)), a non-maximum value can be selected and then over-compensation is suppressed. Note that, unlike average pooling, this regularization does not affect the feature values themselves; it only affects the selection of the value from the kernel.
\par
Fig.~\ref{fig:pooling-effect}(c) shows the result of the regularized pooling operation (with the $3\times 3$ window). The holes of digit images and fine structures of texture images are well preserved even after the pooling operations, compared with the max pooling operation (b). We can thus expect that our regularized pooling can avoid over-compensation and thus keep the separability among classes. It should be noted that this property will lead to a stable training process with a faster convergence because it will be possible to avoid local minima due to the over-compensation.\par

The main contributions of this paper are summarized as follows:
\begin{itemize}
    \item We propose a regularized pooling operation whose capability in terms of deformation compensation fits the characteristics of actual deformations. To the best of authors' knowledge, this is the first proposal of the regularized pooling operation.
    \item Since the regularized pooling operation can avoid over-compensation and thus preserve essential inter-class differences, it has positive effects on both the training and testing steps. We experimentally show these effects; our regularized pooling operation accelerates the training step (i.e., provides quick convergence) and improves the recognition accuracy, especially by avoiding confusion between similar classes, such as `7' and `9' and `a' and `e.'  In a qualitative study, we also observed that the proposed method can preserve important inter-class differences.
    \item We investigate when the proposed method is superior to max pooling using different datasets such as handwritten character image datasets and a texture image dataset. The experiment with texture images also shows the structure preservation capability of the regularized pooling operation.
\end{itemize}

\section{Related Work}
In recent years, many researchers have focused on pooling operations to improve the performance of deep learning-based architectures~\cite{gong2014multi,gao2016compact,husain2019remap}. Pooling operations can reduce the dimension of the input features and render them to invariant to small shifts and deformations~\cite{lecun2015deep}. However, the spatial information lost in the traditional pooling layers causes problems that limit the learning capability of deep neural networks~\cite{feng2011geometric,laptev2016ti}.

\subsection{Traditional pooling operations}
To handle the problems in the traditional pooling operations, many methods have been proposed to extend or improve them in different ways~\cite{graham2014fractional,10.1007/978-3-319-11740-9_34,zhai2017s3pool,DBLP:journals/corr/abs-1804-02702,Wei_2019_CVPR}. To solve the problems that the MP2-pooling ($2\times 2$ max pooling) reduces the size of the hidden layers quickly and the disjointed nature of the regions of pooling can limit generalization, Graham \cite{graham2014fractional} proposed fractional max pooling (FMP) to reduce the size of the image by a factor of $\alpha$ with $1<\alpha<2$. Zhai \textit{et al.} \cite{zhai2017s3pool} proposed S3Pool, which extends standard max pooling by decomposing pooling into two steps: max pooling with stride one and a non-deterministic spatial downsampling step by randomly sampling rows and columns from a feature map. They observed that this general stochasticity acts as a strong regularizer, and can also be seen as performing implicit data augmentation by introducing distortions to the feature maps. To regularize CNN-based architectures, Yu \textit{et al.} \cite{10.1007/978-3-319-11740-9_34} proposed mixed pooling that was inspired by the random dropout \cite{hinton2012improving} and DropConnect \cite{wan2013regularization} methods. Similarly, Wei \textit{et al.} \cite{Wei_2019_CVPR} proposed an intermediate form between max and average pooling called polynomial pooling (P-pooling) to provide an optimally balanced and self-adjusted pooling strategy for semantic segmentation. To compensate for spatial information lost in the max pooling layer, Zheng \textit{et al.} \cite{ZHENG2019558} extracted displacement directions from the max pooling layers and combined them with the original max pooling features to capture structural deformations in text recognition tasks.

\subsection{Recent pooling operations}
Considering the limitations of traditional pooling methods, many pooling operations and layers have recently been proposed to address problems in traditional pooling methods pertaining to specific applications such as image detection and classification~\cite{he2015spatial,saeedan2018detail,kobayashi2019global,gao2019lip}, 
handwriting and text recognition~\cite{graham2014fractional,ZHENG2019558,nguyenvan2019pooling}, semantic segmentation~\cite{bulo2017loss,he2017std2p,Wei_2019_CVPR}, 
and other challenging computer vision tasks~\cite{DBLP:journals/corr/abs-1805-11123,saeedan2018detail,zhang2019local,liu2019simple}. 
He \textit{et al.}~\cite{he2015spatial} introduced a spatial pyramid pooling (SPP) layer to remove the fixed-size constraint on the network, thereby making the network robust to object deformation. 
Kobayashi~\cite{kobayashi2019global} proposed a trainable local pooling function guided by global features beyond local ones. 
The parameterized pooling form is derived from a probabilistic perspective to flexibly represent various types of pooling, and the parameters are estimated by using statistics of the input feature map. More recently, Gao \textit{et al.} \cite{gao2019lip} proposed Local Importance-based Pooling (LIP) that can automatically enhance discriminative features during the downsampling procedure by learning adaptive importance weights based on the inputs. 
LIP solved the problem that the traditional downsampling layers can prevent discriminative details from being well preserved, which is crucial for the recognition and detection tasks.

Compared with prevalent pooling operations, the proposed regularized pooling considers spatial information and regulates the directions of pooling to be homogenized around the neighboring kernels. The advantage of the proposed method is that it compensates for deformations when the neighboring parts shift to random directions. In this way, the proposed method becomes more effective than conventional pooling methods at accelerating convergence.

\begin{figure}[!t]
	\centering
	\includegraphics[width=0.6\hsize]{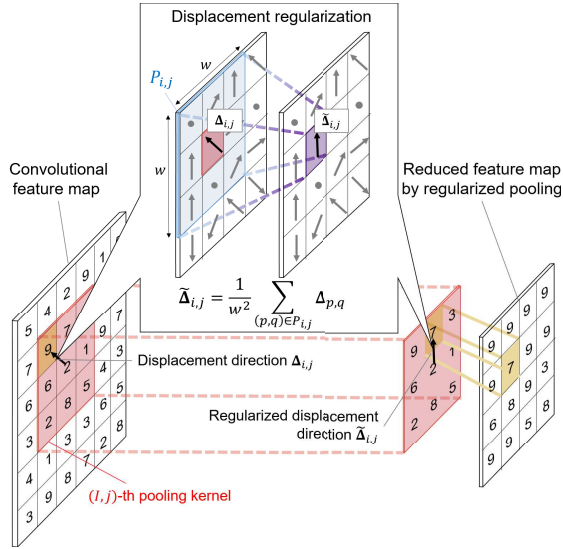}
	\caption{Overview of the regularized pooling operation, where the pooling kernel size $n = 3$, smoothing window size $w = 3$, and stride $s = 1$. The direction from the center of the kernel to the maximum value, called the displacement direction $\bm{\Delta}_{i,j}$, is first calculated for the $(i,j)$-th kernel. For example, in the red area of the figure, the direction from `2' to `9' is treated as the displacement direction $\bm{\Delta}_{i,j}$. Displacement directions for the entire feature map are then regularized via smoothing. As shown in the blue area, the averaged displacement directions around the target pixel is treated as the regularized displacement direction $\widetilde{\bm{\Delta}}_{i,j}$. Finally, the reduced feature map is calculated by extracting the pixel value in the kernel indicated by the direction of the regularized displacement direction.}
	\label{fig:apply}\vspace{-4mm}
\end{figure}

\section{Regularized Pooling}
Fig.~\ref{fig:apply} shows an overview of regularized pooling. Regularized pooling takes a convolutional feature map as its input and outputs a new feature map. Although the outline of the calculation is similar to that of max pooling, the main difference is that the direction to the maximum value in a kernel, called the displacement direction, is extracted and then revised by the displacement directions at the neighboring kernels. 

Specifically, the displacement direction is first extracted from the input feature map by the max pooling operation. Assume that we can conduct the max pooling operations $I$ times vertically and $J$ times horizontally by sliding an $n \times n$ kernel with the stride $s$~\footnote{To be specific, given a convolutional feature map of size $H \times W$ as input, $I = \lfloor (H - 1)/s \rfloor + 1$ and $J = \lfloor (W - 1)/s \rfloor + 1$ if we add a proper size of padding to the input.}. 
For the $(i,j)$-th pooling kernel ($i\in\{1,\ldots,I\}, j\in\{1,\ldots,J\}$), the displacement direction $\bm{\Delta}_{i,j}$ is defined as the direction from the center of the kernel to the maximum value. 
The possible value for the element of $\bm{\Delta}_{i,j}$, $(\bm{\Delta}_{i,j})_k$ $(k = 1, 2)$, depends on the parity of $n$.    
For an odd $n$, $(\bm{\Delta}_{i,j})_k \in \{-\frac{n - 1}{2}, \ldots, -1, 0, 1, \ldots, \frac{n - 1}{2} \}$, whereas $(\bm{\Delta}_{i,j})_k \in \{-\frac{n}{2}, \ldots, -1, 1, \ldots, \frac{n}{2} \}$ for an even $n$. 
The displacement directions are then regularized by considering the adjacent displacement directions. The regularization is based on spatial smoothing of the displacement directions. The regularized displacement direction $\widetilde{\bm{\Delta}}_{i,j}$ is calculated as follows: \begin{eqnarray}
  \widetilde{\bm{\Delta}}_{i,j} = \frac{1}{w^2}\sum_{(p, q) \in P_{ij}}{\bm{\Delta}_{p,q}}, 
  \label{eq:regularization}
\end{eqnarray}
where the odd integer $w$ is the size of the smoothing window and $P_{ij} = \{\bm{\Delta}_{p,q} | p \in \{i-\frac{w-1}{2}, \ldots, i+\frac{w-1}{2}\}, q \in \{j-\frac{w-1}{2}, \ldots, j+\frac{w-1}{2}\} \}$. 
Finally, the output feature map is generated by using the regularized displacement directions. The pixel value in the $(i,j)$-th kernel indicated by the regularized displacement direction $\widetilde{\bm{\Delta}}_{i,j}$ is extracted as the $(i,j)$-th value of the reduced feature map. 

Note that $\widetilde{\bm{\Delta}}_{i,j}$ can be a non-integer vector due to the smoothing in Eq. (\ref{eq:regularization}) while it should be an integer vector for the acquisition of a reduced feature map. Therefore,  we quantize $\widetilde{\bm{\Delta}}_{i,j}$ if it is a non-integer. For an odd $n$, the element of $\widetilde{\bm{\Delta}}_{i,j}$ is rounded to the nearest integer \footnote{If the fraction part is exactly 0.5, it is rounded away from zero.}. 
For an even $n$, the element is rounded away from zero, so as not to be zero.

\vspace{-3.5mm}
\section{Experiment on Character Images}
\label{section:experiment1}
We first assessed the effectiveness of the regularized pooling
operation by comparing it with traditional pooling operations. In particular, we verified that regularized pooling improves the convergence speed of learning through a comparison of performance profiles. Second, we qualitatively show that regularized pooling reduces the dimensionality of the input feature map while preserving detailed structures via example-based evaluation. Finally, we evaluate the effects of the kernel size, smoothing window size, and stride, which are important hyperparameters of regularized pooling.
\vspace{-3.5mm}
\subsection{Dataset}
We evaluated our regularized pooling on two standard benchmark datasets of handwritten character images, MNIST~\cite{lecun1998gradient} and EMNIST~\cite{DBLP:journals/corr/CohenATS17}. Character images often undergo various and severe deformations; however, those deformations are still continuous and topology-preserving so as not to spoil inter-class differences. Therefore, character images are the most suitable for understanding the characteristics of the proposed regularized pooling operation. 
MNIST is comprised of $28 \times 28$ handwritten digit images and split to  $60,000$ training samples and $10,000$ test samples.  EMNIST is comprised of uppercase and lowercase English alphabet letters with 37 classes (after several identifications between indistinguishable classes, such as `o' and `O') and $88,800$ for training and $14,800$ for test. \\

\vspace{-5.5mm}
\subsection{Experimental setup}
The network architecture used in this experiment is summarized in Table \ref{table:network}. In the table, ``conv, $3 \times 3$, $64$'' represents a convolutional layer with a $3 \times 3$-sized $64$-channel kernel. This network was based on VGG~\cite{Simonyan15} with some convolutional blocks and fully connected layers removed to fit the network to the size of the input image. Two convolutional layers with a ReLU activation function were cascaded as a block, and a pooling layer was connected after the convolutional block. After repeating this convolutional and pooling connection three times, a fully-connected (FC) layer with a softmax activation was connected as the last layer. Dropout with a ratio of $0.25$ was used for the last FC layer. Regularized pooling was applied to the first pooling layer. For comparison, we used max pooling and average pooling. 

In all experiments, we calculated the average of five trials by changing the initial weights of the network when computing classification accuracy. To clarify the effect of pooling, all images were resized to $60\times60$. Zero-padding was not used in any pooling operation. We used the SGD optimizer for weight updating. The learning rate was $10^{-2}$ for MNIST and $10^{-4}$ for EMNIST. The number of learning epochs and the batch size were set to $50$ and $100$, respectively. We employed cross entropy as a loss function. 

\begin{figure}[t]
\centering
\includegraphics[width=0.31\textwidth]{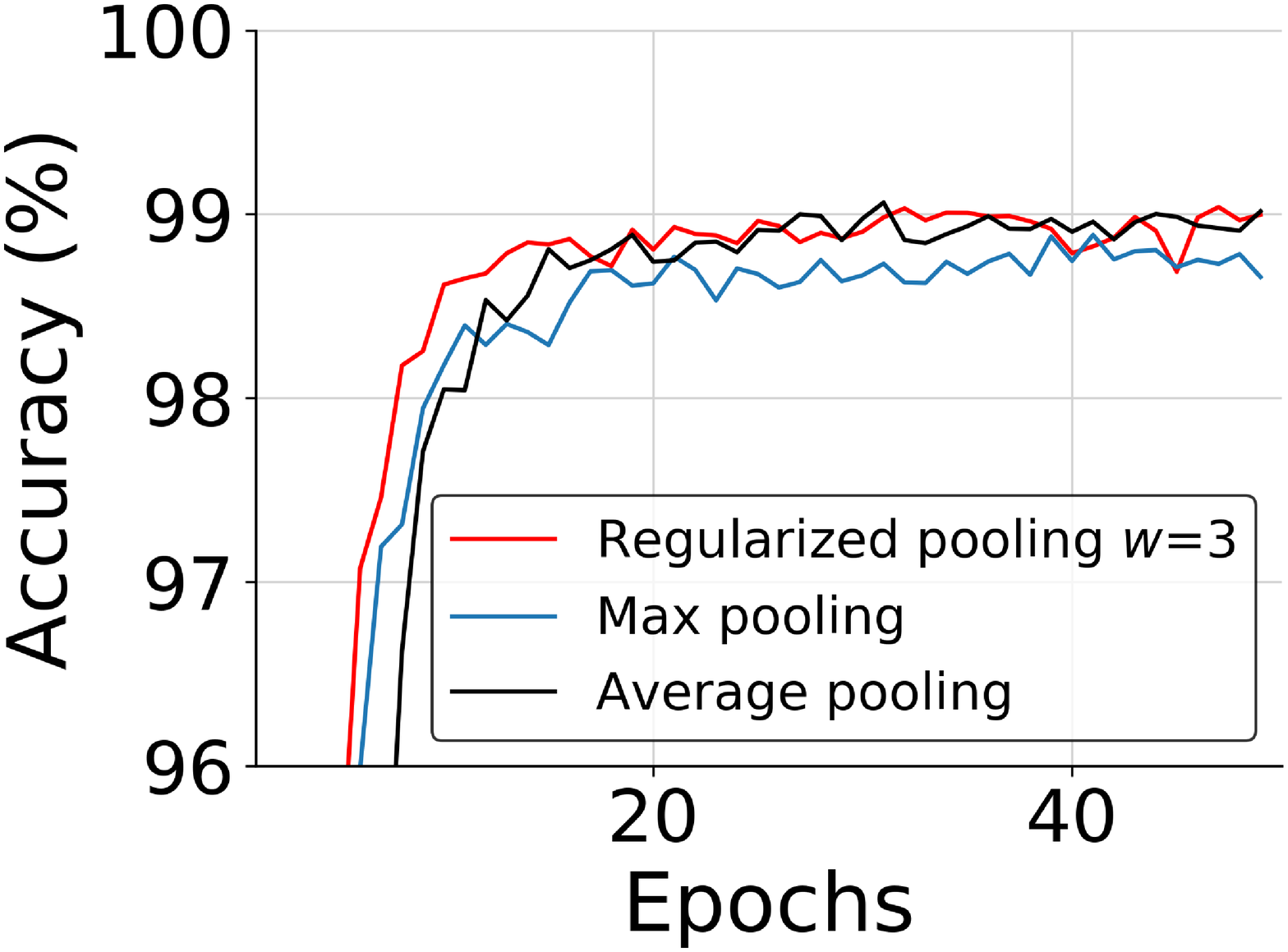}
\qquad
\includegraphics[width=0.31\textwidth]{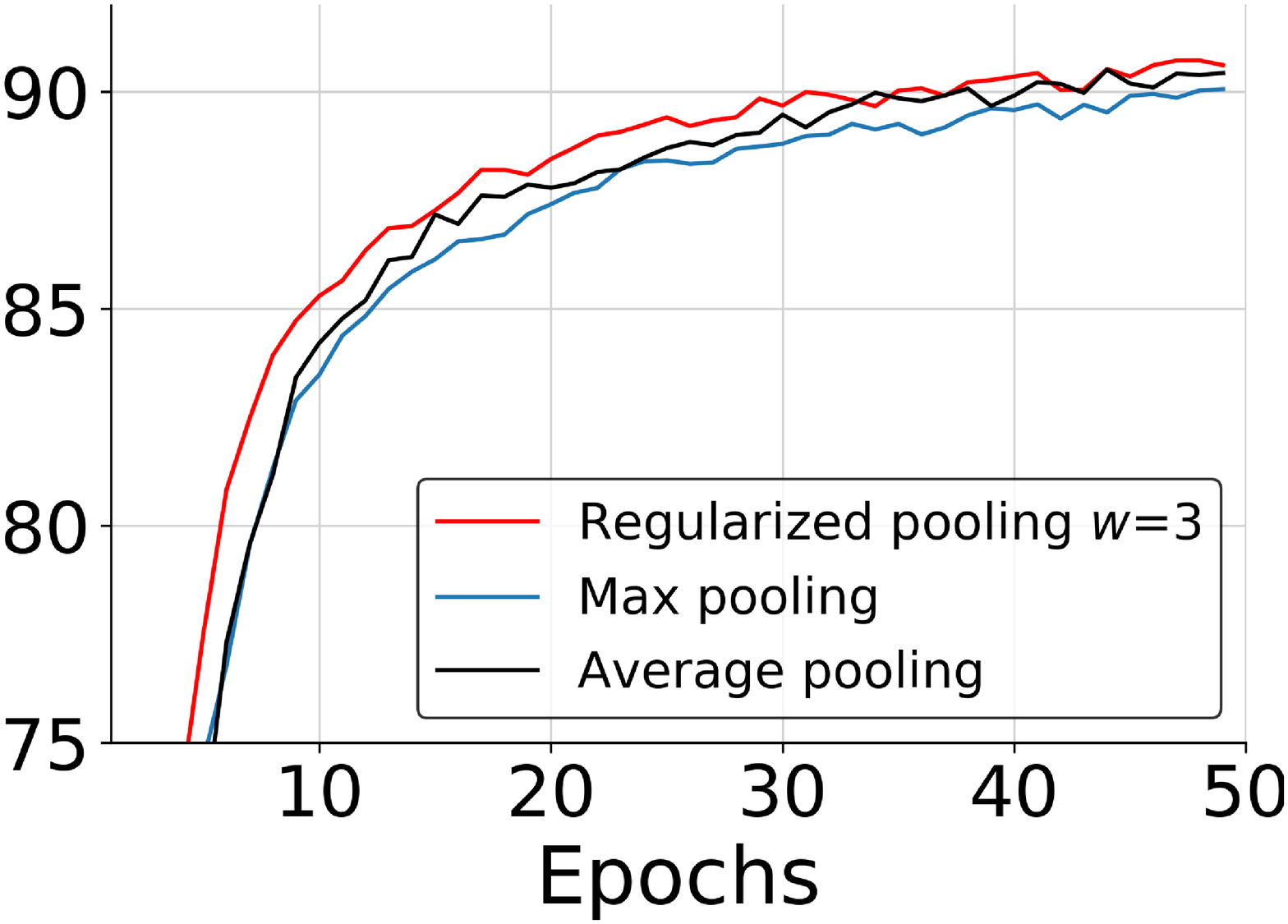}
\\[-4mm]
\caption{Comparison of performance profiles among regularized pooling, max pooling, and average pooling on the MNIST (left) and EMNIST (right) datasets.}
\label{fig:acc_comparison}\vspace{-2mm}
\end{figure}
\begin{table}[t]
    \begin{minipage}{0.48\textwidth}
    \caption{Network architecture}
    \centering
    \begin{footnotesize}
    \begin{tabular}{|c|c|c|}
    \hline
    Name                   & Output                        & Layer \\ \hline
    \multirow{2}{*}{conv1} & \multirow{2}{*}{$60\times60$} & conv, $3\times3$, $64$ \\
                          &                               & conv, $3\times3$, $64$ \\ \hline
    \multirow{3}{*}{pool1} & \multirow{5}{*}{$12\times12$} & \textbf{regularized pool}, $n \times n$, or \\
                          &                               & \textbf{max pool}, $n \times n$, or\\
                          &                               & \textbf{average pool}, $n \times n$\\ \cline{1-1} \cline{3-3}
    \multirow{2}{*}{conv2} &                               & conv, $3\times3$, $128$ \\
                          &                               & conv, $3\times3$, $128$ \\ \hline
    pool2                  & \multirow{3}{*}{$6\times6$}   & max pool, $2\times2$ \\ \cline{1-1} \cline{3-3}
    \multirow{2}{*}{conv3} &                               & conv, $3\times3$, $256$ \\ 
                          &                               & conv, $3\times3$, $256$ \\ \hline
    pool3                  & $3\times3$                    & max pool, $2\times2$ \\ \hline 
    FC                     & $3\times3$                    & FC + softmax \\ \hline
    \end{tabular}
    \label{table:network}
    \end{footnotesize}
    \end{minipage}
    \makeatletter
    \def\@captype{figure}
    \makeatother
	\begin{minipage}{0.48\textwidth}
	\centering
	\includegraphics[width=0.75\hsize]{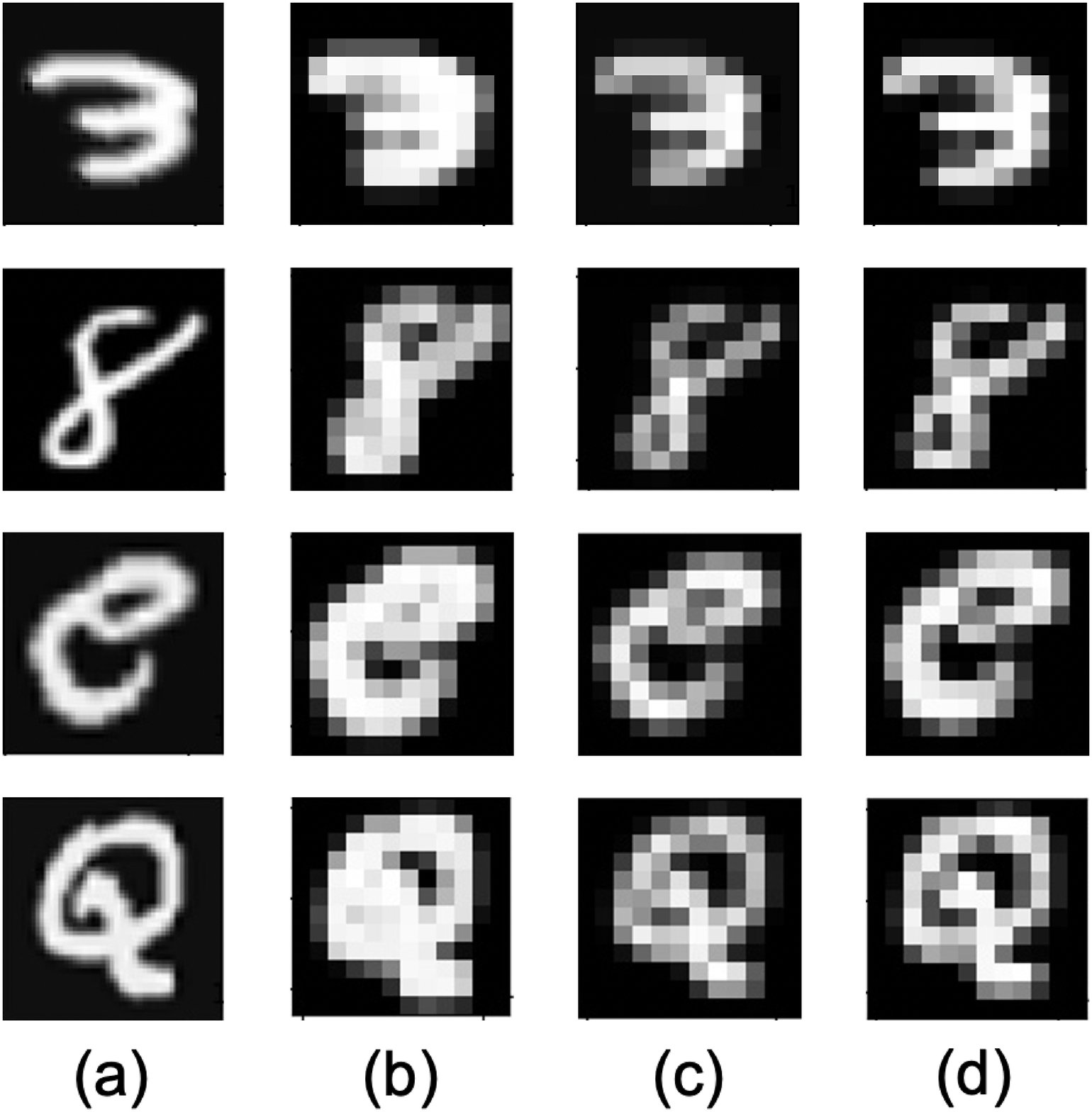}
	\caption{Visualization of the feature maps after the pooling operation. (a)~Original feature map. (b)~Max pooling. (c)~Average pooling. (d)~Regularized pooling.}
	\label{fig:visualization}
	\end{minipage}\vspace{-3mm}
\end{table}

\vspace{-2.5mm}
\subsection{Performance comparison with traditional pooling methods}
Fig.~\ref{fig:acc_comparison} shows the comparison of performance profiles among regularized pooling, max pooling, and average pooling on the test datasets of MNIST and EMNIST. In this figure, the pooling kernel size $n$, smoothing window size $w$, and stride were set to $n = 5$, $w = 3$, and $s = 5$, respectively. Note that every line shows the average of five trials by changing the initial weights of the network.\par
These results confirmed that the learning convergence of regularized pooling is faster than those of max pooling and average pooling. 
Compared to max pooling, our regularized pooling could suppress the excessive deformation compensations and thus could avoid local minima due to them, especially the early training stages, when the feature values tend to have random-like values and the deformation compensation ability of max pooling is abused. Examples that support the above hypotheses are provided in the next subsection.\par
It is also very important that regularized pooling is better than average pooling. Regularized pooling still keeps important (large) feature values compared to average pooling. This is because feature values themselves are smoothed by average pooling, whereas they are not smoothed by our regularized pooling---regularized pooling just smooths the selection direction. 
 
\vspace{-1.5mm}
\subsection{Qualitative evaluation}
We qualitatively evaluated the differences between regularized pooling and traditional pooling methods by visualizing the feature maps after the application of the pooling operations. The visualization examples are shown in Fig.~\ref{fig:visualization}. In max pooling, the shapes of the characters collapsed due to over-compensation. For example, the holes of '8' and 'Q' are filled with white pixels. In average pooling, the outlines of the characters are blurred although their shapes are preserved better than by max pooling. This is because average pooling considered the surrounding information by smoothing the feature values directly. Conversely, regularized pooling preserved both the shapes and the outlines of the characters better than max pooling and average pooling because it considers surrounding information by regularizing the deformation features, without directly smoothing the input feature maps. 

\begin{figure}[!t]
    \centering
    \subfigure[`7' and `9']{
    \includegraphics[width=0.31\hsize]{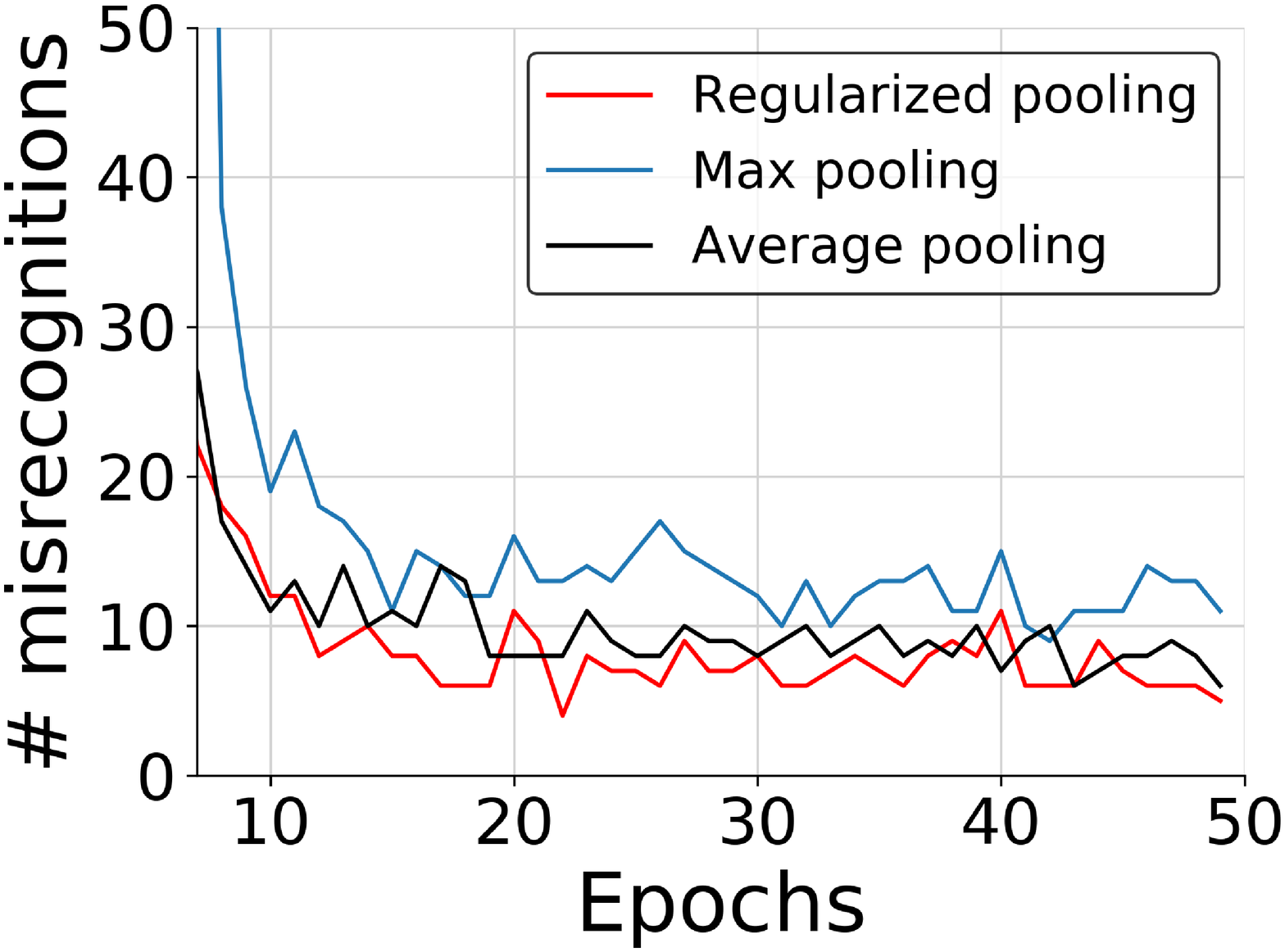}
    \label{fig:mis1}}
    \subfigure[`a' and `e']{
    \includegraphics[width=0.31\hsize]{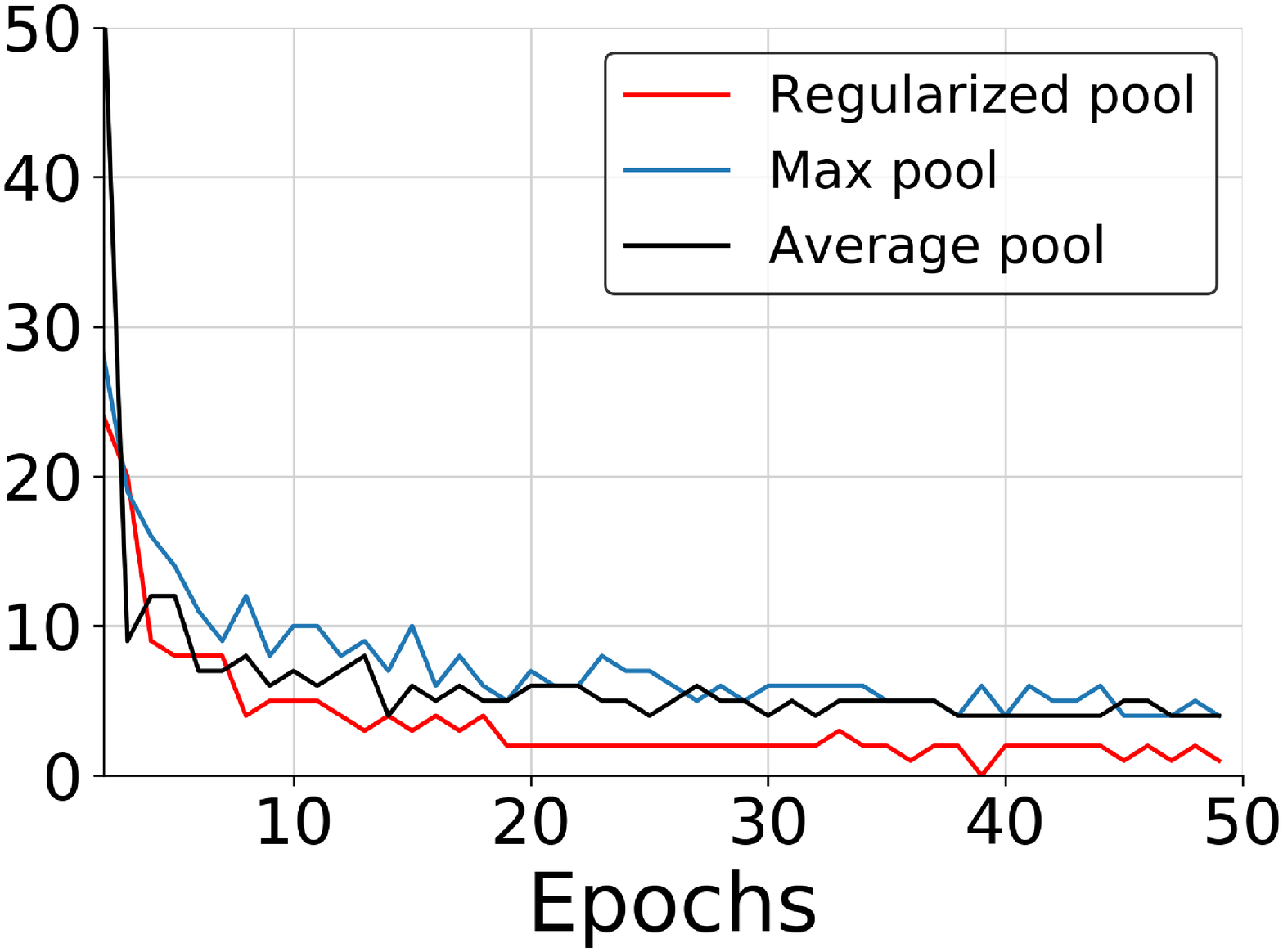}
    \label{fig:mis3}}\\[-3mm]
    \subfigure[`2' and `7']{
    \includegraphics[width=0.31\hsize]{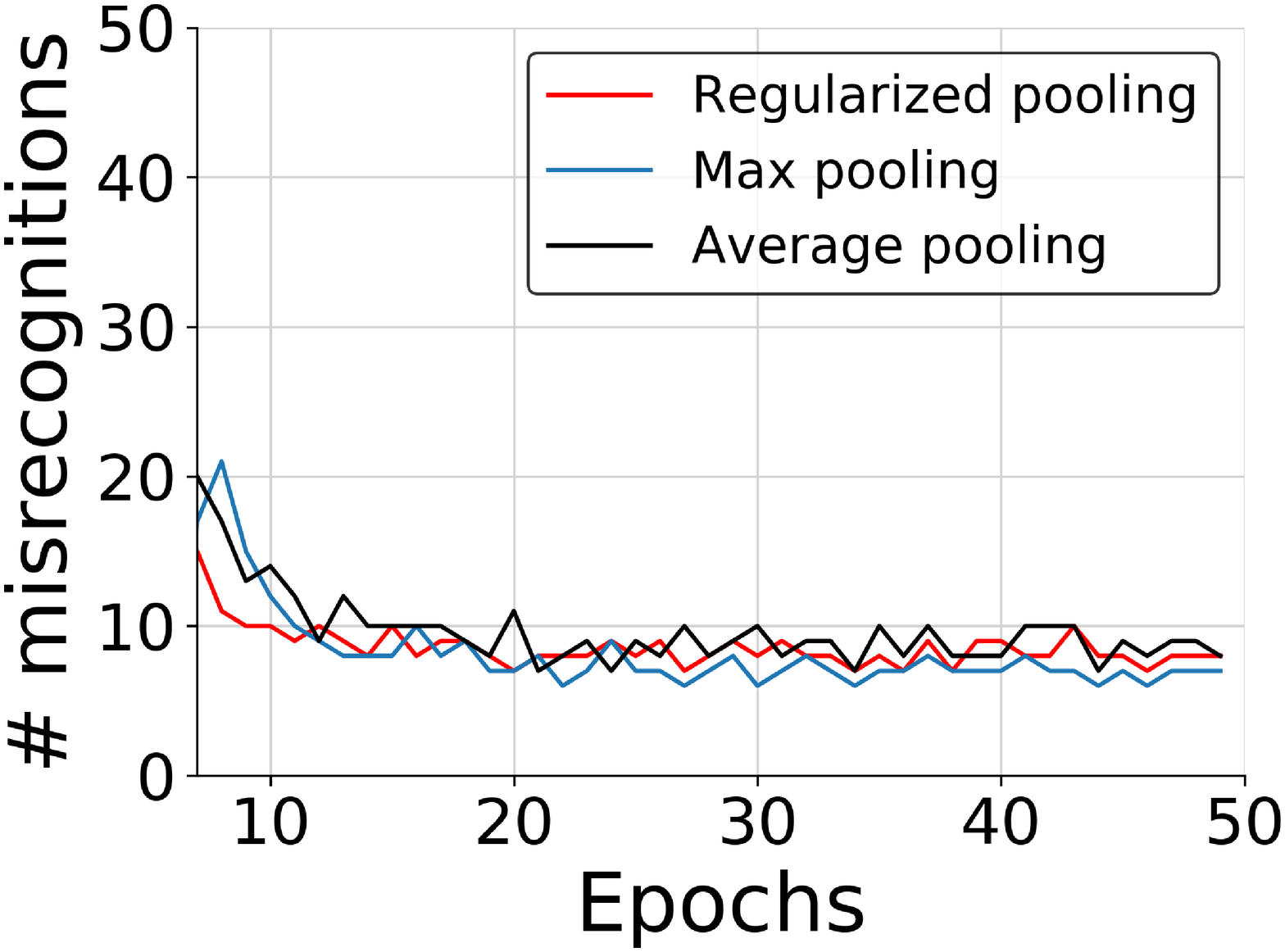}
    \label{fig:mis2}}
    \subfigure[`C' and `O']{
    \includegraphics[width=0.31\hsize]{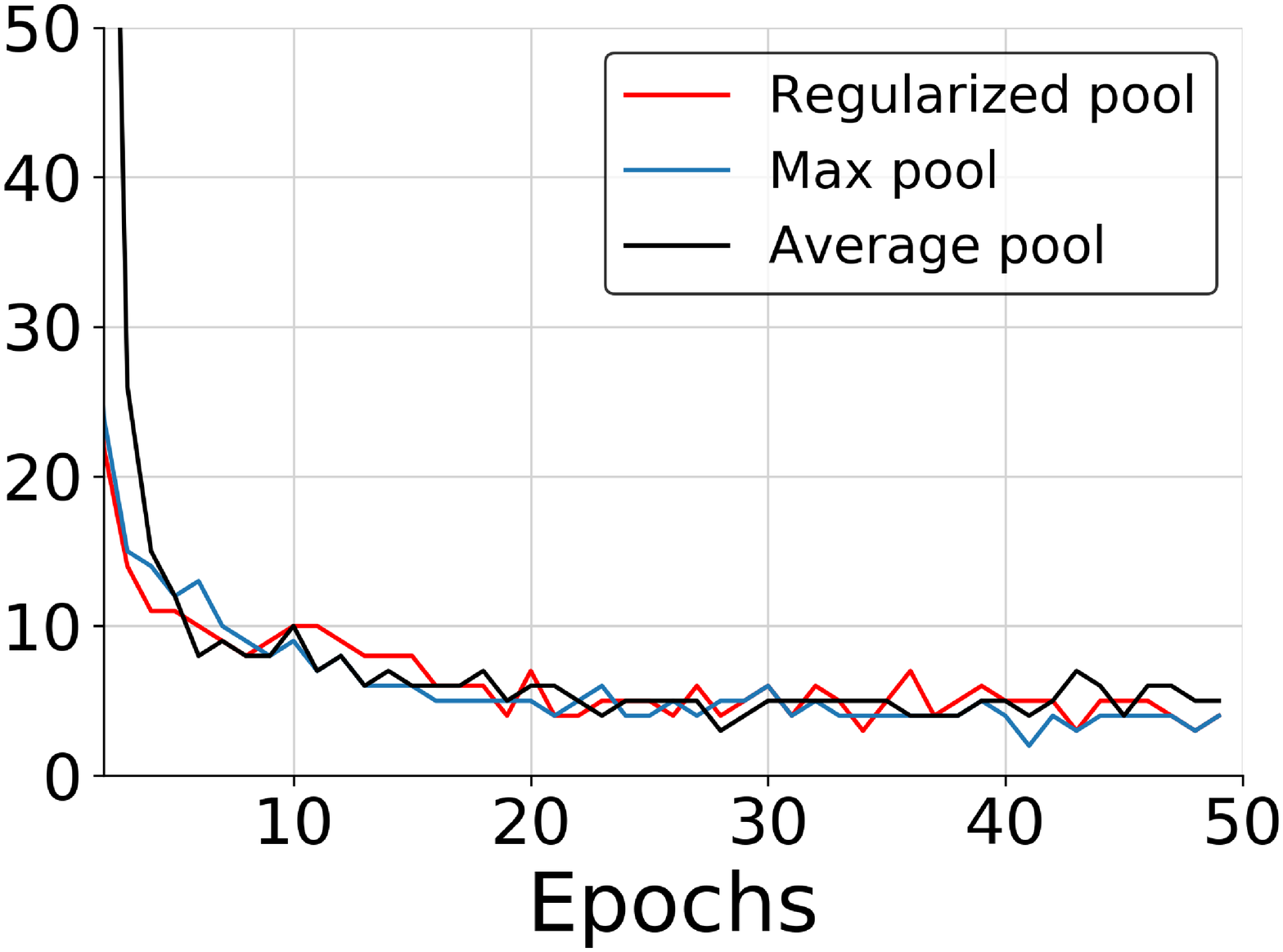}
    \label{fig:mis4}}\\[-4mm]
    \caption{The number of misrecognitions between specific class pairs. (a) and (b)~are confusing pairs, and (c) and (d)~are easy pairs.}
    \label{fig:misrecognition}\vspace{-3mm}
\end{figure}

We verified how the qualitative differences among the pooling methods in the above visualization affected recognition errors. Fig.~\ref{fig:misrecognition} shows the number of misrecognitions between certain class pairs along with the learning epochs. In Figs.~\ref{fig:mis1} and \ref{fig:mis3}, `7' and `9,' and `a' and `e,' are given as the pairs whose structural differences are subtle, i.e., confusing pairs. In addition, Figs.~\ref{fig:mis2} and \ref{fig:mis4} show the misrecognitions between the pairs of `2' and `7,' and `C' and `O,' where there are clear structural differences in the handwritten images, i.e., easy pairs. For the confusing pairs, regularized pooling reduced  misrecognitions compared with max pooling and average pooling, whereas there was no remarkable difference among the three pooling methods for the easy pairs. These results show that regularized pooling preserves the detailed structure of the input feature map by suppressing over-compensations and thus effectively distinguishes between class pairs with subtle structural differences.

\vspace{-1.5mm}
\subsection{Effect of hyperparameters}
\begin{figure}[t]
    \centering
    \subfigure[$n = 3$ (MNIST)]{
    \includegraphics[width=0.23\hsize]{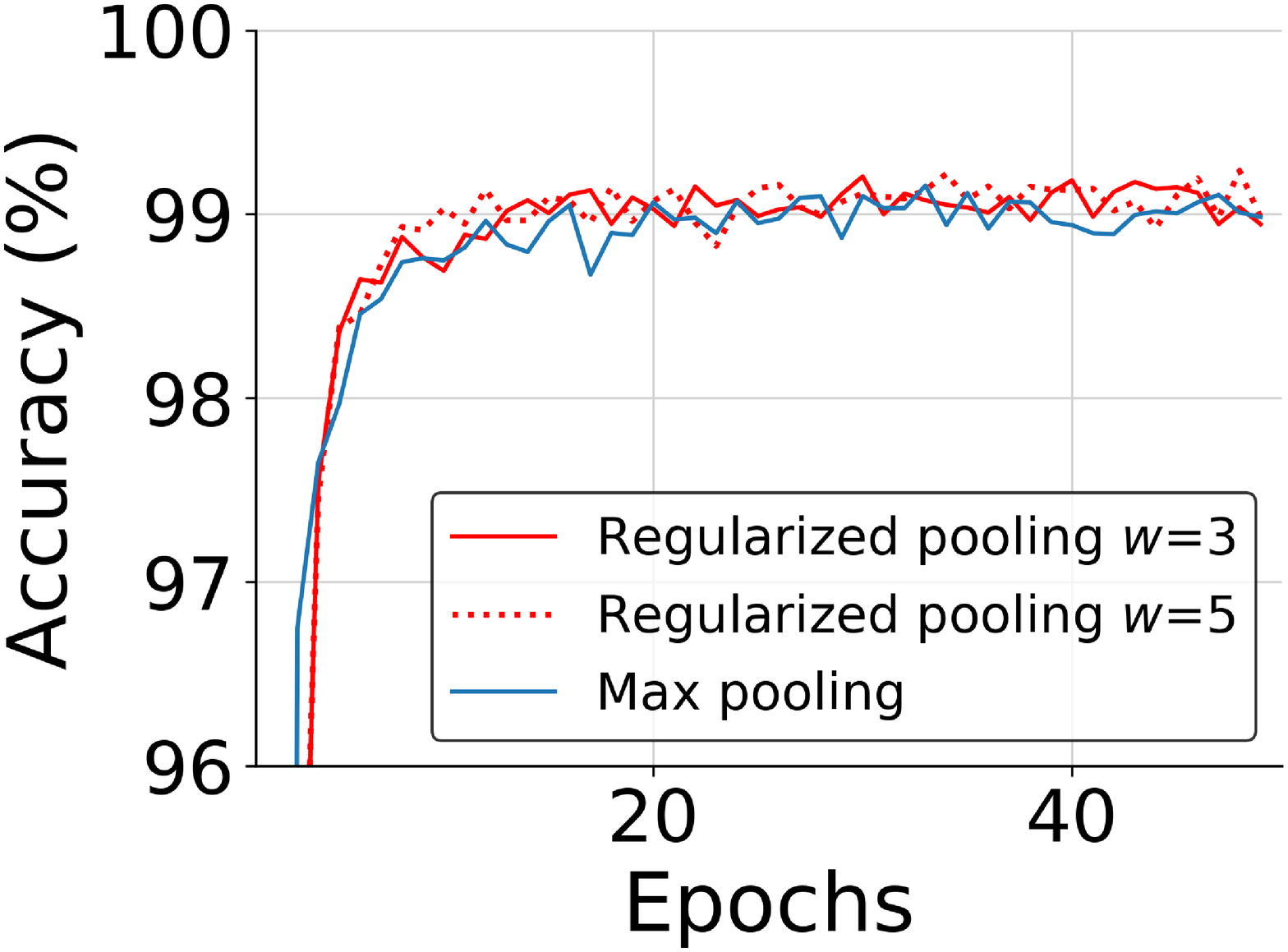}
    \label{fig:result1}}
    \subfigure[$n = 5$ (MNIST)]{
    \includegraphics[width=0.23\hsize]{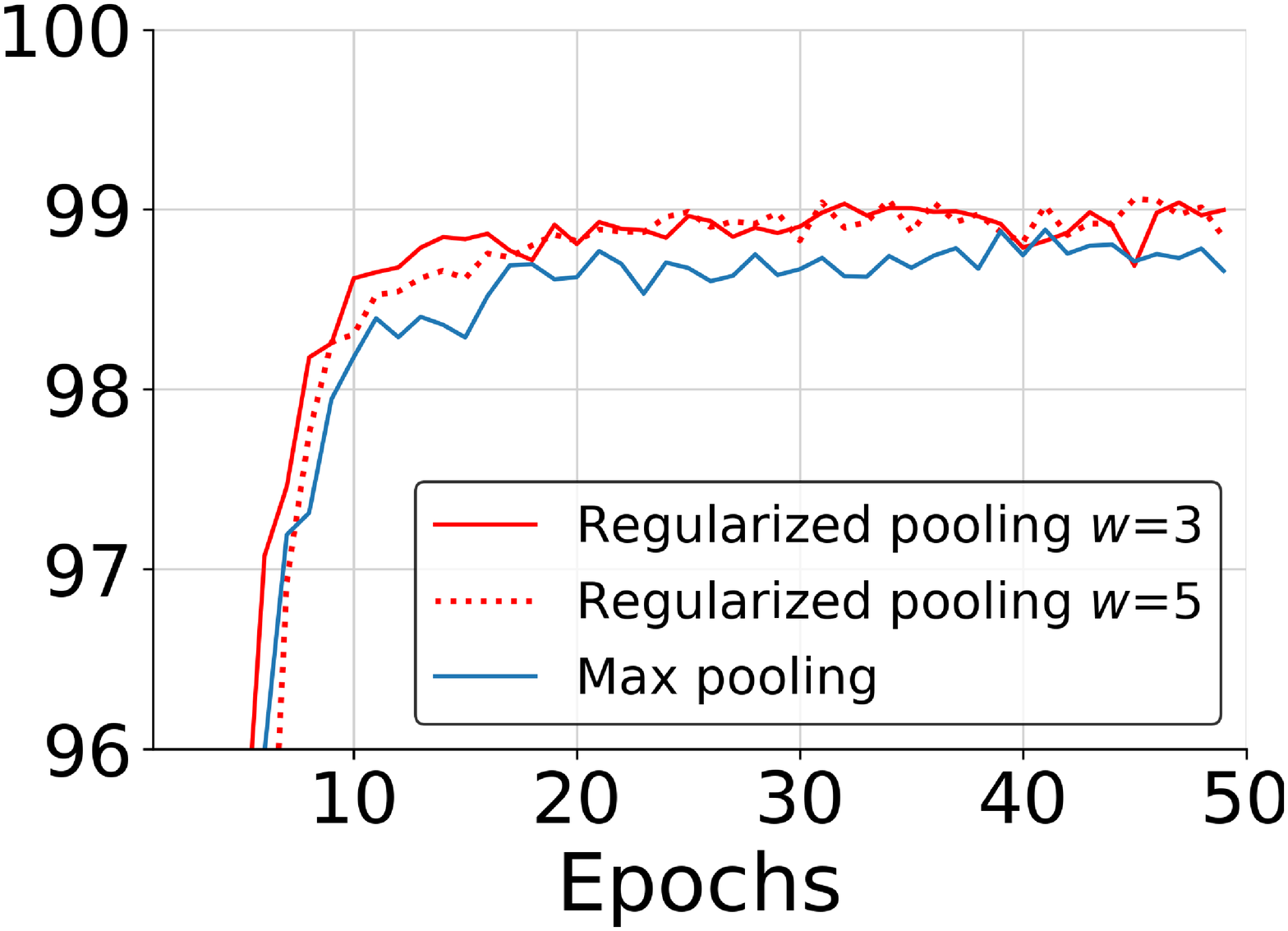}
    \label{fig:result2}}
    \subfigure[$n = 3$ (EMNIST)]{
    \includegraphics[width=0.23\hsize]{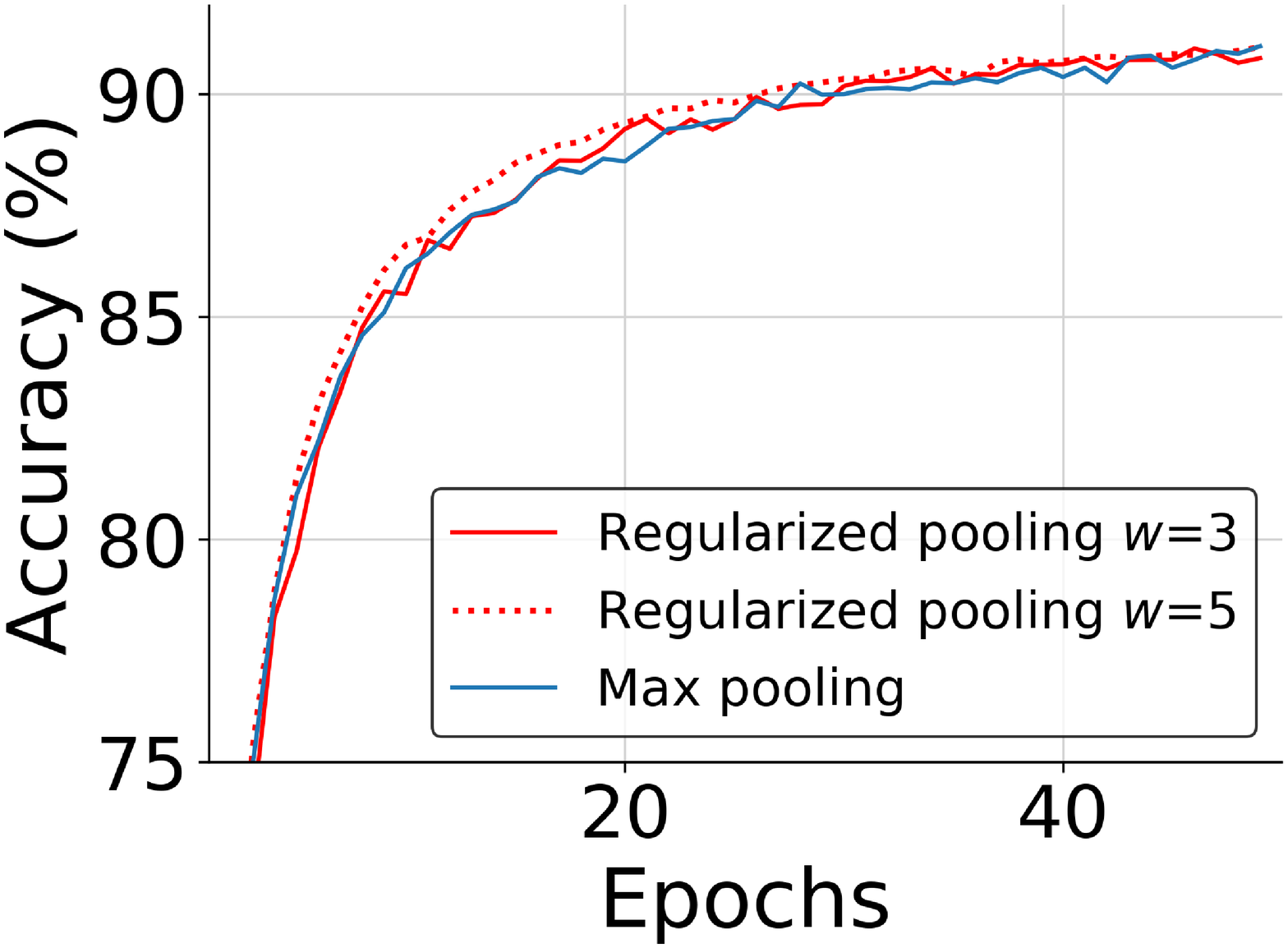}
    \label{fig:result3}}
    \subfigure[$n = 5$ (EMNIST)]{
    \includegraphics[width=0.23\hsize]{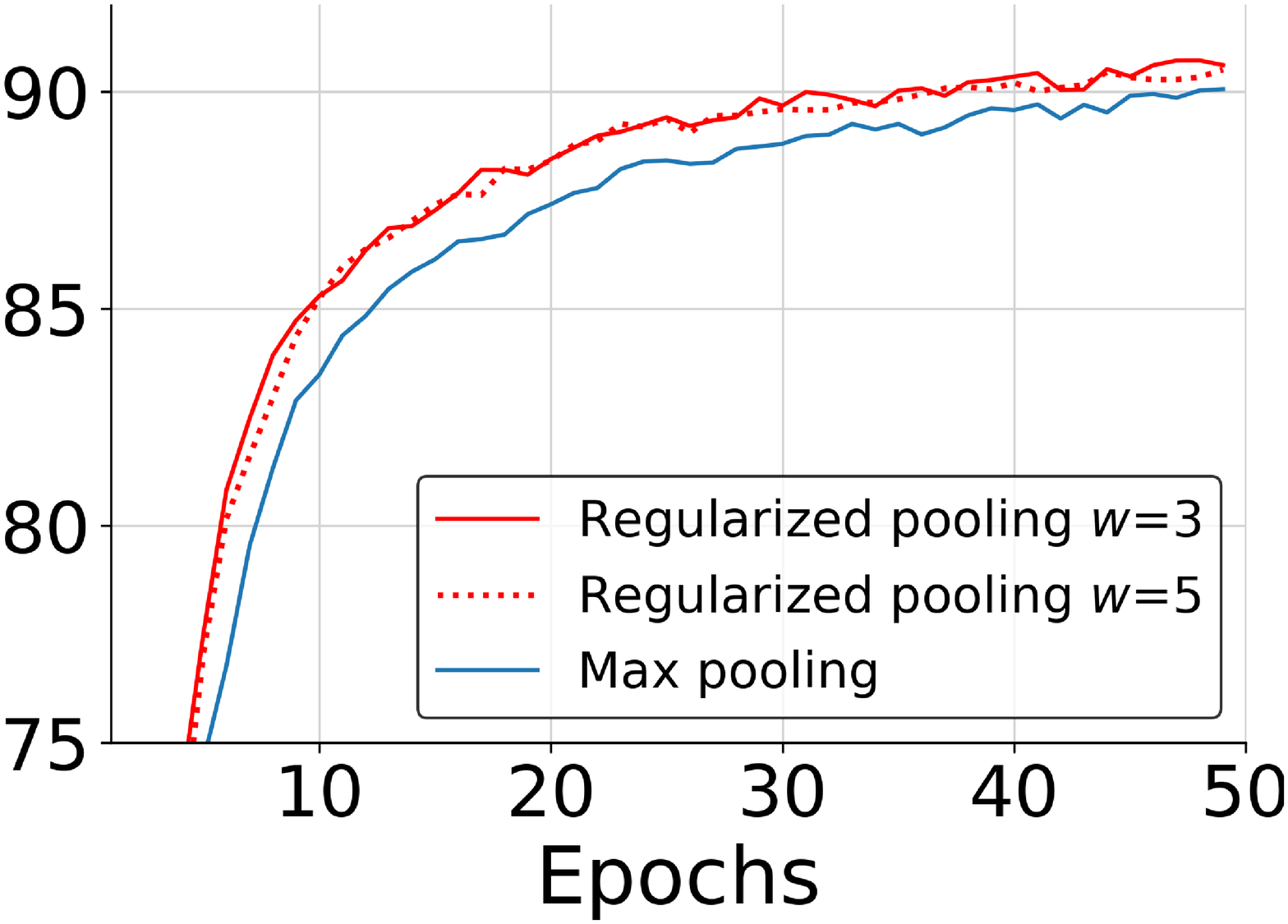}
    \label{fig:result4}}\\[-4mm]
    \caption{Effects of the pooling kernel size $n$ and smoothing window size $w$.}
    \label{fig:result}
    \centering
    \subfigure[Regularized pooling]{
    \includegraphics[scale=0.12]{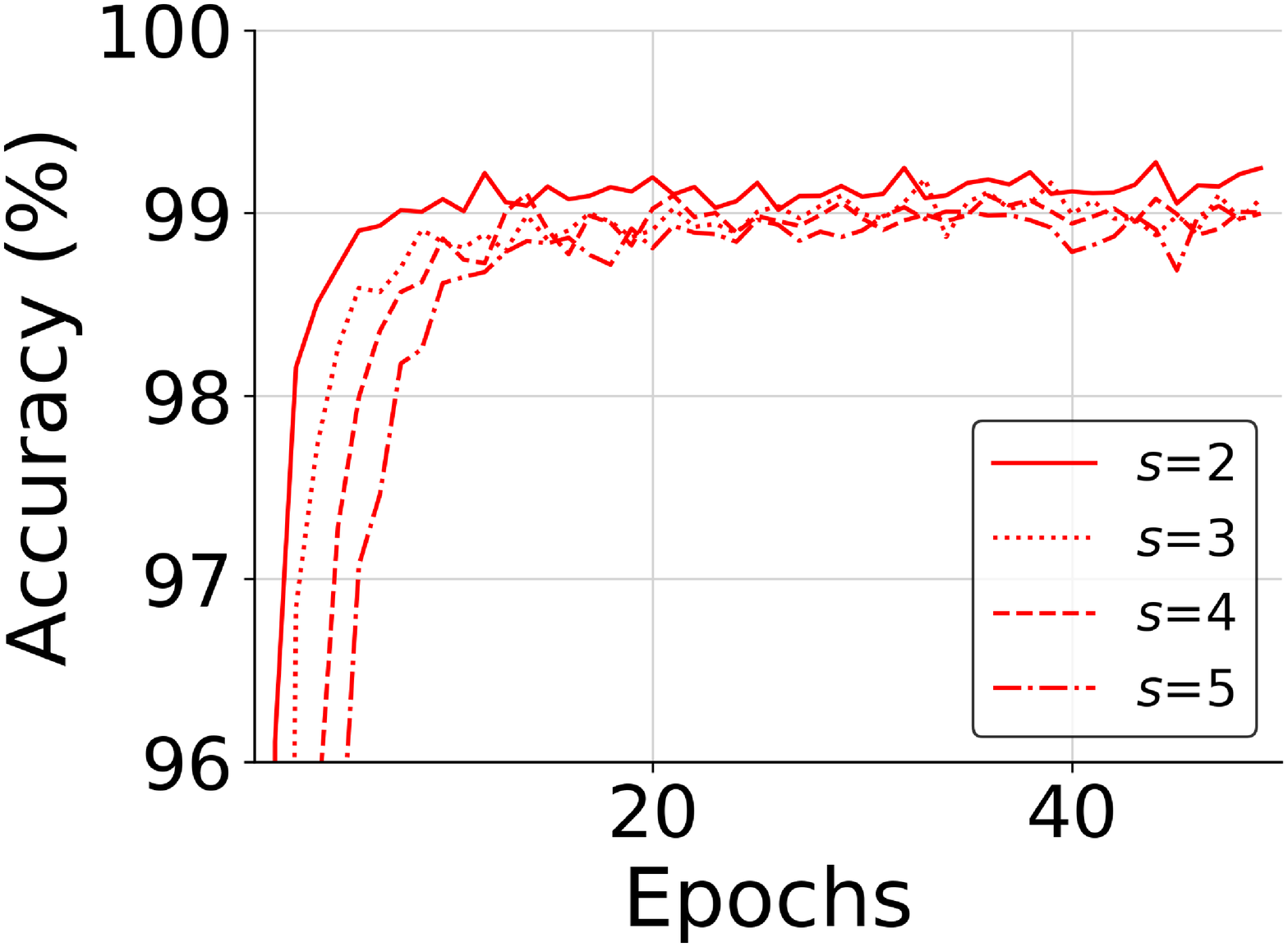}
    \label{fig:overlap1}}
    \subfigure[Max pooling]{
    \includegraphics[scale=0.12]{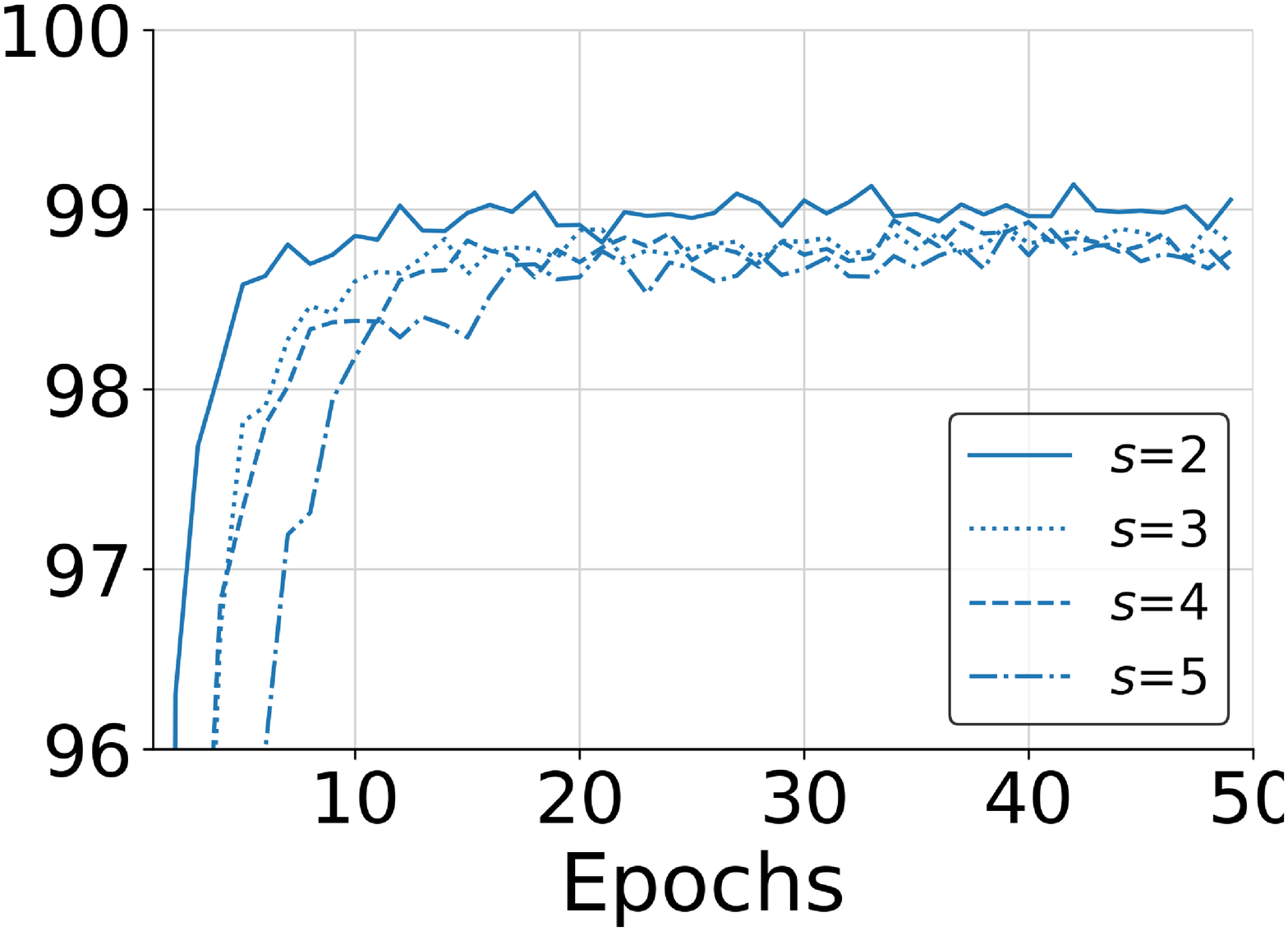}
    \label{fig:overlap2}}\\[-4mm]
    \caption{Performance profiles when varying the stride $s$.}
    \label{fig:overlap}
    \subfigure[10 epochs]{
    \includegraphics[width=0.48\hsize]{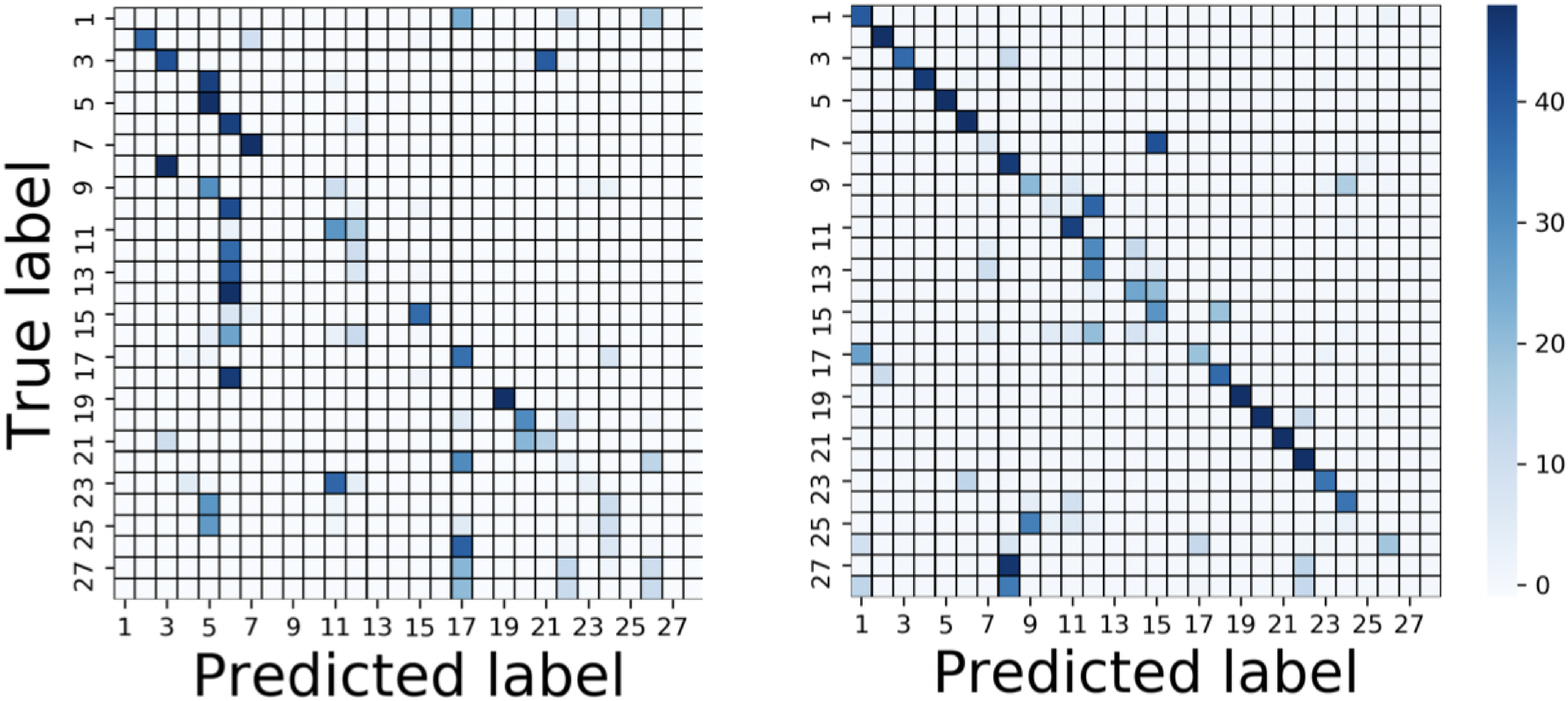}
    \label{fig:confmat1}}
    \subfigure[40 epochs]{
    \includegraphics[width=0.48\hsize]{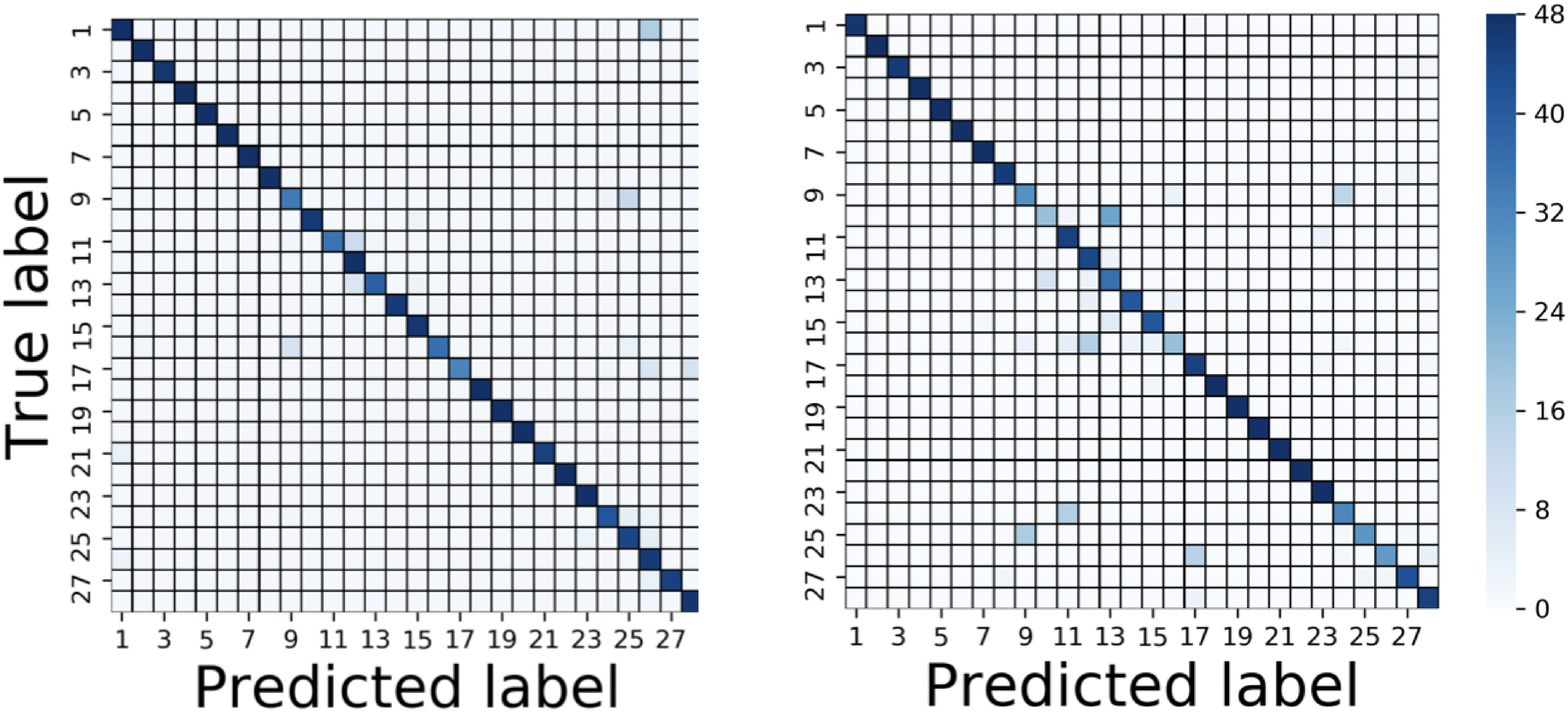}
    \label{fig:confmat2}}\\[-4mm]
    \caption{Confusion matrix of texture recognition. Left: max pooling. Right: regularized pooling.}
    \label{fig:texture-cm}\vspace{-3mm}
\end{figure}

We evaluated the effect of the hyperparameters, i.e., the pooling kernel size $n$, smoothing window size $w$, and stride $s$. Fig.~\ref{fig:result} shows the performance profiles when $n$ and $w$ were varied to $n = 3$ and $5$ and $w=3$ and $5$. The results by max pooling are also shown for comparison. These results suggest that the effect of $n$ on the results is more significant than $w$. Moreover, the difference between regularized pooling and max pooling was clearer when $n$ was larger. This is because the larger the value of $n$ was, the stronger the effect of over-compensation due to max pooling was, whereas regularized pooling suppressed it. 

The effect of the stride $s$ is shown in Fig.~\ref{fig:overlap}. This figure summarizes the performance profiles of regularized pooling and max pooling on the MNIST dataset while $s$ was varied to $s = 2, 3, 4,$ and $5$. 
The result shows that regularized pooling showed faster convergence than max pooling at all $s$ values, while a smaller stride $s$ yielded better performance.  


\section{Experiment on Texture Images}
In this experiment, we aimed to clarify the characteristics of regularized pooling by analyzing the results of classification for texture images with various structures. In particular, we reveal the kind of images for which regularized pooling is effective. 

We used the Kylberg texture dataset \cite{Kylberg2011c} that contains $28$ classes with $160$ unique samples each ($112\times28=5376$ samples for training, $48\times28=1344$ samples for testing). Each sample is a grayscale image of size $576 \times 576$ pixels. We resized all images to $256 \times 256$ in the experiment. For weight updating, we used the Adam optimizer with parameters of $10^{-4}$, $\beta_1=0.9$, and $\beta_2 = 0.99$. The batch size was set to $32$. The network architecture and other experimental conditions were the same as in the experiments described in Section~\ref{section:experiment1}. 

\begin{figure}[!t]
    \centering
    \includegraphics[width=0.6\hsize]{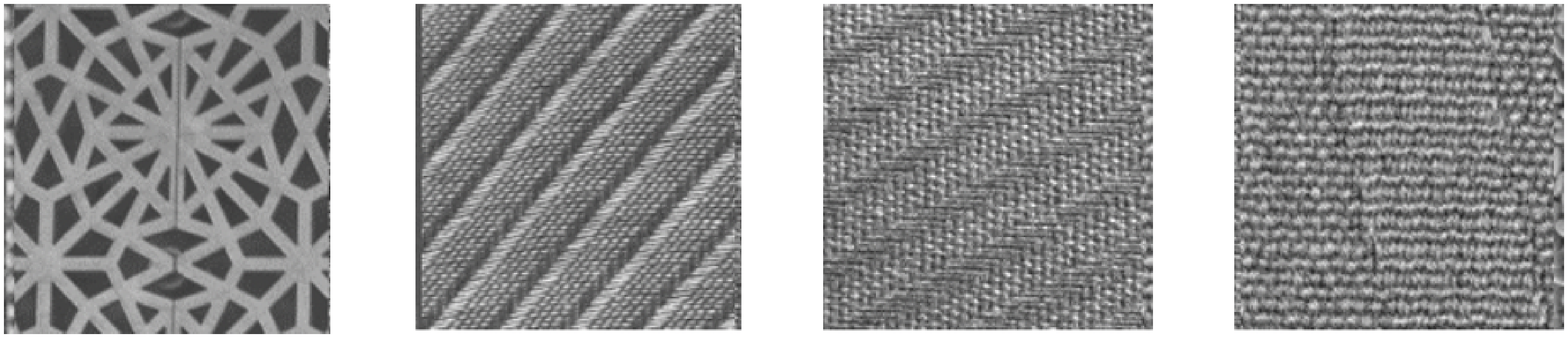}\\[-2mm]
    (a)~Improved classes, 6, 19, 20, and 21.
    \includegraphics[width=0.6\hsize]{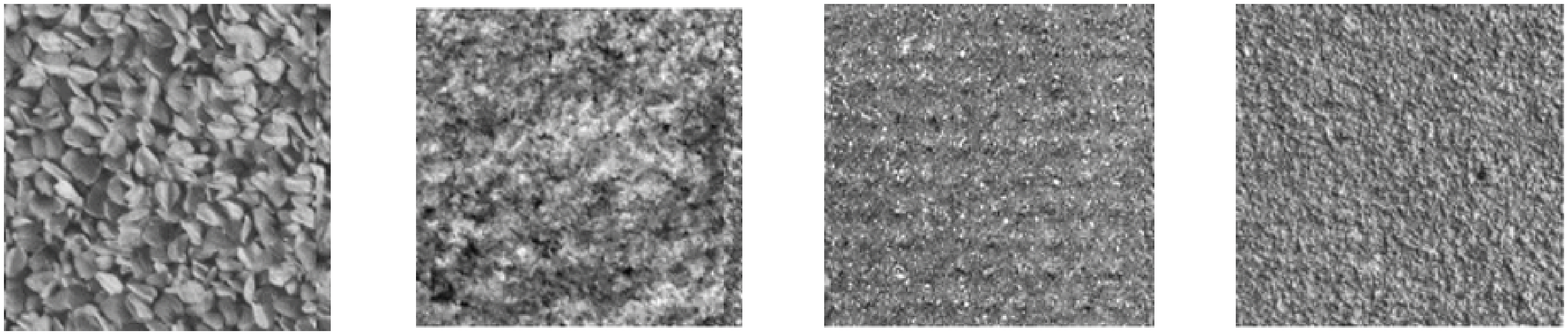}\\[-2mm]
    (b)~Degraded classes, 10, 23, 26, and 27.\\[-3mm]
    \caption{Examples texture images from the improved and degraded classes.}
    \label{fig:texture-ex}\vspace{-5mm}
\end{figure}

Fig.~\ref{fig:texture-cm} shows the confusion matrix on the test set obtained by using max pooling and regularized pooling at 10 and 40 epochs. According to Fig.~\ref{fig:texture-cm}(a), certain classes such as class 6, 19, 20, and 21 are almost completely correctly recognized in the early stage of learning. Example images from the improved classes by regularized pooling are shown in Fig.~\ref{fig:texture-ex}(a). 
The common feature of these images was that they had a periodic structure. Regularized pooling could retain this periodic structure to some extent and thus show superiority.
In Fig.~\ref{fig:texture-cm}(b), however, several classes, such as class 10, 23, 26, and 27, were not correctly recognized by regularized pooling, even at the 40 epoch. Example images from these classes are shown in Fig.~\ref{fig:texture-ex}(b), and it can be seen that they are near-random patterns without any specific periodicity, i.e., no clear structure. These results demonstrated that regularized pooling is effective for patterns with a periodic structure. This is because regularized pooling performs spatially continuous operations between adjacent kernels, and therefore preserves frequency information to some extent in the feature map after pooling.

\vspace{-1.0mm}
\section{Conclusion}
We proposed regularized pooling, which enables a local pooling operation suitable for actual deformations. In the traditional max pooling operation, the value selection direction is determined as the maximum value position at each kernel independently. By considering it as a deformation compensation process, this independent strategy will cause over-compensation. In contrast, our regularized pooling operation smooths the value selection directions over the neighboring kernels to suppress over-compensation and thus stabilizes the training process. Through experiments on image recognition, we demonstrated that regularized pooling improves separability of similar classes and the convergence of learning compared with the conventional pooling methods. \par

In future work, we will further consider another strategy for smoothing the value selection directions, although we have shown that even simple average-based smoothing is already effective. For example, using an adaptive window size controlled by some spatial and/or channel-wise attention mechanisms will be a possible choice.

\vspace{-1.0mm}
\section*{Acknowledgments}
This work was supported by JSPS KAKENHI Grant Number JP17H06100 and JST ACT-I Grant Number JPMJPR18UO.
\vspace{-1.0mm}
\bibliography{main}
\bibliographystyle{splncs04}
\end{document}